\documentclass[11pt]{article}

\usepackage[utf8]{inputenc}
\usepackage[T1]{fontenc}
\usepackage{amsmath,amssymb,amsthm}
\usepackage{graphicx}
\usepackage{booktabs}
\usepackage{array}
\usepackage{longtable}
\usepackage{geometry}
\usepackage[colorlinks=true,linkcolor=blue,citecolor=blue,urlcolor=blue]{hyperref}
\usepackage{natbib}
\usepackage{enumitem}
\usepackage{xcolor}
\usepackage{caption}
\usepackage{float}

\newtheorem{assumption}{Assumption}

\title{\textbf{F-DACE: Fuzzy Disagreement-Aware Causal Evidence Fusion for Abstention-Safe Conversational Retail Decision Support}}

\author{
Sourish Dey\\
Data Science and Machine Learning, Centric Software\\
\texttt{sourish.dey@centricsoftware.com}, \texttt{sourish.syntel@gmail.com}\\
ORCID: 0009-0002-3565-7304
}

\date{29.06.2026}

\begin{document}

\maketitle

\begin{center}
\small Corresponding author: Sourish Dey. Correspondence: \texttt{sourish.dey@centricsoftware.com} and \texttt{sourish.syntel@gmail.com}.
\end{center}

\begin{abstract}
Observational decision-support systems often expose one causal estimate as a recommendation even when plausible estimators disagree. The inherent engine of the proposed system is causal machine learning: a conditional-average-treatment-effect estimand identified by backdoor adjustment, estimated by an EconML DML causal forest and DoWhy linear regression, checked by two-way fixed effects, and converted into candidate levers by constrained optimisation. F-DACE is the decision layer on that engine. It represents precision, propensity overlap, placebo-refutation stability, interval overlap, and directional agreement as fuzzy memberships. Hard vetoes force abstention after estimand mismatch, failed diagnostics, informative sign conflict, or weak evidence. In 180 panel simulations spanning six identification conditions, F-DACE made a decision in 67.2\% of runs and limited false recommendations to 17.2\%; the corresponding rates were 33.3\% for the causal forest and 35.6\% for backdoor regression, matching deterministic unanimity rather than dominating it. Nearly all (30 of 31) false recommendations occurred under shared unmeasured confounding, which no fusion rule can diagnose when every component shares the omitted variable. The retail application aggregates a public Walmart panel to 6,435 store-weeks across 45 stores. F-DACE abstains for all five markdown indicators: some estimates are imprecise, one refutation fails, and MarkDown5 has a direct sign conflict. A LangGraph conversational agent exposes impact, what-if, and lever-optimization tools while a deterministic verifier preserves causal-layer status. On 24 live questions it achieved 100.0\% tool-routing accuracy, 100.0\% status fidelity, and 0.983 mean groundedness. On ten adversarial questions it resisted all injected instructions. The system therefore couples a proposed causal decision gate with an empirically evaluated conversational interface.
\end{abstract}

\noindent\textbf{Keywords:} fuzzy evidence fusion; causal machine learning; selective decision making; heterogeneous treatment effects; tool-using agents; conversational decision support; retail analytics

\section{Introduction}

Business analytics increasingly answers interventional questions: whether a promotion should be activated, which lever should be changed, and when a targeted action is preferable to no action. Historical prediction is insufficient for these questions because actions are assigned in response to expected demand. A model may therefore attribute holiday or seasonal demand to a markdown. Causal inference makes the identifying assumptions explicit, but it does not remove the practical problem that different defensible estimators can return incompatible answers from the same observational panel.

Most analytical interfaces hide that incompatibility. A point estimate is selected, formatted, and delivered; uncertainty becomes a parenthetical qualification. This is especially hazardous when a language model is used as the interface, because linguistic fluency can turn a fragile estimate into a persuasive recommendation. A conversational agent remains valuable because it translates managerial intent, maintains scope over multiple turns, selects an analytical operation, and explains its result. The methodological problem is to make that interface preserve rather than erase causal uncertainty: heterogeneous evidence must become a decision or an explicit abstention that fluent answer synthesis cannot override.

The inherent engine of the system is causal machine learning, not language generation. Potential outcomes define the estimand. Unconfoundedness and overlap justify backdoor adjustment on an explicit graph. A double/debiased-machine-learning causal forest estimates how that effect varies across contexts and supplies the leaf-level slopes used for constrained lever search. A DoWhy model identifies the same contrast by backdoor linear regression and stress-tests it with placebo, random-common-cause, and subset refuters. A two-way fixed-effects specification absorbs unit and common-time shocks as a third check. These estimators remain the source of every number the tools return.

F-DACE sits on that engine as a selective decision layer. It does not replace identification, residualization, or refutation. It consumes aligned estimates, standard errors, intervals, overlap, and placebo stability; represents them as fuzzy memberships; and converts disagreement into recommend, do-not-recommend, or mandatory abstention. A veto prevents compensation: a precise forest cannot override a failed overlap diagnostic or an informative sign conflict. The conversational agent then routes managerial questions to the engine and is forbidden from rewriting an abstention as advice.

The contributions are:
\begin{itemize}[leftmargin=*]
\item F-DACE, a fuzzy disagreement-aware fusion policy that converts aligned causal evidence---precision, propensity overlap, refutation stability, interval overlap, and directional agreement---into a recommendation, a do-not-recommend decision, or a mandatory abstention, with an explicit estimand contract and non-compensatory vetoes.
\item A decision-risk evaluation protocol that scores false recommendations, decision coverage, selective error, and regret across six seeded identification conditions, rather than reporting estimator fit alone.
\item A store--week retail application in which apparently positive component estimates still yield a justified no-recommendation output, together with the boundary case that shared unmeasured confounding remains undiagnosable by any fusion rule.
\item A fail-closed conversational decision-support layer in which a causal abstention cannot be rewritten as advice, evaluated end-to-end for tool routing, answer grounding, causal-status fidelity, multi-turn scope continuity, a deterministic non-LLM baseline, and adversarial instruction resistance.
\end{itemize}

\section{Related work}

\subsection{Causal machine learning and panel estimators}

Potential-outcome and graphical frameworks define effects by interventions rather than observed associations \citep{pearl2009,imbens2015}. Causal and generalized random forests estimate conditional average treatment effects with adaptive partitions and valid intervals \citep{wager2018,athey2019}. Double/debiased machine learning supplies the orthogonal scores and cross-fitting that let flexible nuisance learners be used without first-order bias \citep{chernozhukov2018,foster2023}. The residual-on-residual reduction of \citet{robinson1988} and the R-learner objective \citep{nie2021} are the estimating equations of the forest used here, implemented as EconML CausalForestDML \citep{battocchi2019}. Backdoor adjustment makes the assumed confounder set explicit; DoWhy operationalizes model, identify, estimate, and refute \citep{sharma2020}. Two-way fixed effects are common panel checks, although repeated treatment and heterogeneous timing can invalidate a simple coefficient interpretation \citep{dechaisemartin2020,goodmanbacon2021}. These methods answer different failure risks. They are therefore retained as distinct engines and fused only after their estimands are aligned.

\subsection{Fuzzy evidence and abstention}

Fuzzy sets represent graded membership rather than forcing uncertain evidence into crisp categories \citep{zadeh1965}. In decision systems, this permits precision, validity, and agreement to remain separate signals before aggregation. Selective prediction adds a reject option when expected error is too high \citep{chow1970,geifman2017}. F-DACE combines these ideas for causal evidence: memberships summarize partial support, while non-compensatory vetoes preserve identification diagnostics. It differs from model averaging because it does not average causal effects into a new effect estimate.

\subsection{Conversational causal systems}

ReAct interleaves language-model reasoning with calls to external tools, while Toolformer and graph-based orchestration demonstrate how models can select and sequence structured operations \citep{yao2023,schick2023,langchain2024}. Recent causal systems include PrecAIse, Causal-Copilot, and CausalAgent \citep{orderique2024,wang2025,zhu2026}. They establish the value of conversational access to causal workflows. The open systems question is whether answer synthesis preserves a causal engine's decision status when the engine refuses to recommend.

This work treats the agent as a second research layer rather than a decorative front end. Tool routing tests semantic interpretation; multi-turn state tests scope continuity; groundedness tests whether claims originate in tool outputs; and status fidelity tests whether abstention or failure survives generation. Automatic judges are retained as diagnostic signals, not correctness oracles, because a relevant answer can still be based on an invalid estimate and a safe refusal can be scored as irrelevant to a request for certainty \citep{zheng2023,es2024}.

\section{Causal machine learning engine}

F-DACE is a decision policy over causal estimates. Those estimates are produced by a fixed engine whose objects are the conditional average treatment effect, backdoor identification, double machine learning, constrained lever search, and a two-way panel check. This section states that engine before the fusion rule.

\subsection{Estimand and identifying assumptions}

Let $Y$ denote weekly sales, $T$ a treatment (a markdown or other promotional lever), and $X$ a covariate vector that includes holiday status, fuel price, CPI, unemployment, temperature, seasonality, and store characteristics. Write $Y(t)$ for the potential outcome under intervention $T = t$ \citep{imbens2015}. The target of the heterogeneous-effect engine is the conditional average treatment effect (CATE), with the average treatment effect (ATE) as its population mean:
\begin{equation}
\tau(x) = \mathbb{E}[Y(1) - Y(0) \mid X = x], \qquad \text{ATE} = \mathbb{E}[\tau(X)]
\end{equation}

$\tau(x)$ is a CATE at a covariate profile, not the unobservable unit-level counterfactual $Y_i(1) - Y_i(0)$. The causal forest's per-observation output $\hat\tau(X_i)$ is therefore the CATE evaluated at unit $i$'s covariates. Identification from observational data uses two standard assumptions.

\begin{assumption}[Unconfoundedness]
$\{Y(0), Y(1)\} \perp T \mid X$: conditional on $X$, treatment assignment is as good as random.
\end{assumption}

\begin{assumption}[Overlap]
$0 < P(T = 1 \mid X = x) < 1$ for all $x$ in the support of $X$.
\end{assumption}

Under these conditions $\tau(x)$ is identified and the backdoor adjustment holds:
\begin{equation}
\mathbb{E}[Y \mid do(T = t)] = \mathbb{E}_X\big[\, \mathbb{E}[Y \mid T = t, X] \,\big]
\end{equation}

\begin{figure}[H]
\centering
\includegraphics[width=0.6\textwidth]{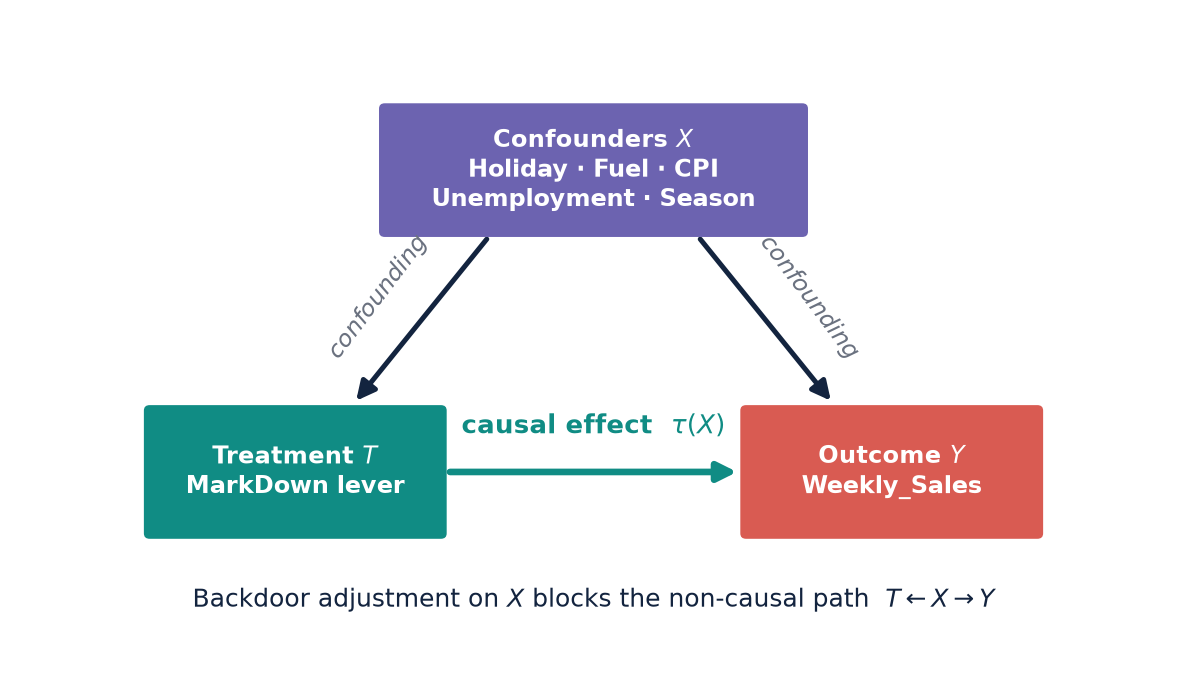}
\caption{Assumed causal graph of the engine. Backdoor adjustment on $X$ blocks the spurious path $T \leftarrow X \rightarrow Y$.}
\label{fig:causal-graph}
\end{figure}

Figure~\ref{fig:causal-graph} shows the assumed graph. Confounders $X$ open the non-causal path $T \leftarrow X \rightarrow Y$; adjustment on $X$ blocks that path so that the remaining $T \rightarrow Y$ edge carries the causal effect $\tau(X)$.

\subsection{Graphical identification and refutation (DoWhy)}

For attribution questions the engine fits a DoWhy CausalModel \citep{sharma2020}. A driver is binarised at its median to form $T$, an explicit graph encodes treatment, outcome, and confounders, the effect is identified by the backdoor criterion, and estimated by \texttt{backdoor.linear\_regression}: the coefficient on $T$ in the ordinary-least-squares regression of $Y$ on $T$ and the adjustment set. The estimate is then stress-tested with three refuters that become F-DACE's validity scores:
\begin{itemize}[leftmargin=*]
\item \textbf{Placebo treatment}: replace $T$ with a random permutation. A valid effect should collapse toward zero.
\item \textbf{Random common cause}: add an independent noise covariate to the adjustment set. A robust effect should be essentially unchanged.
\item \textbf{Data subset}: re-estimate on a random subsample; large swings indicate instability.
\end{itemize}

The same identified linear contrast, stored as a per-unit causal beta, prices what-if queries: impact $\approx \beta \Delta$. That object is the engine's answer to a specified change. It is not yet a recommendation.

\subsection{Double machine learning and the causal forest}

Where the effect varies, the engine uses the partially linear model
\begin{equation}
Y = \tau(X)\, T + g(X) + \varepsilon, \qquad T = m(X) + \eta
\end{equation}
with $\mathbb{E}[\varepsilon \mid X, T] = 0$ and $\mathbb{E}[\eta \mid X] = 0$. Here $g(X)$ is a baseline response surface and $m(X) = \mathbb{E}[T \mid X]$ is the propensity. Defining the outcome nuisance $\ell(X) = \mathbb{E}[Y \mid X]$ and the residuals $\tilde{Y} = Y - \ell(X)$, $\tilde{T} = T - m(X)$, the \citet{robinson1988} decomposition reduces the model to $\tilde{Y} = \tau(X)\tilde{T} + \varepsilon$, whose population solution is the local residual-on-residual regression
\begin{equation}
\tau^*(x) = \arg\min_f \, \mathbb{E}\big[(\tilde{Y} - f(X)\tilde{T})^2 \mid X = x\big] = \mathbb{E}[\tilde{Y}\tilde{T} \mid X = x] \,/\, \mathbb{E}[\tilde{T}^2 \mid X = x]
\end{equation}

The empirical-loss form is the R-learner objective \citep{nie2021}. The estimator is built on the Neyman-orthogonal moment
\begin{equation}
\psi(W; \tau, \eta) = (\tilde{Y} - \tau \tilde{T})\, \tilde{T}, \qquad \eta = (\ell, m)
\end{equation}
which satisfies $\partial_\eta \mathbb{E}[\psi(W; \tau_0, \eta_0)] = 0$ at the truth. Orthogonality means small errors in the learned nuisances have only second-order impact on $\tau$ \citep{chernozhukov2018,foster2023}, so random-forest nuisance models may be used without contaminating the causal estimate. Cross-fitting trains nuisances on the complement of each fold. A causal forest then produces adaptive kernel weights $\alpha_i(x)$ and solves the locally weighted moment
\begin{equation}
\sum_i \alpha_i(x)\, \psi(W_i; \tau(x), \eta) = 0
\end{equation}
yielding observation-level estimates $\hat\tau(X_i)$ and variance estimates for intervals \citep{athey2019,wager2018}. Concretely the engine uses EconML CausalForestDML with random-forest nuisance models \citep{battocchi2019}.

\begin{figure}[H]
\centering
\includegraphics[width=0.75\textwidth]{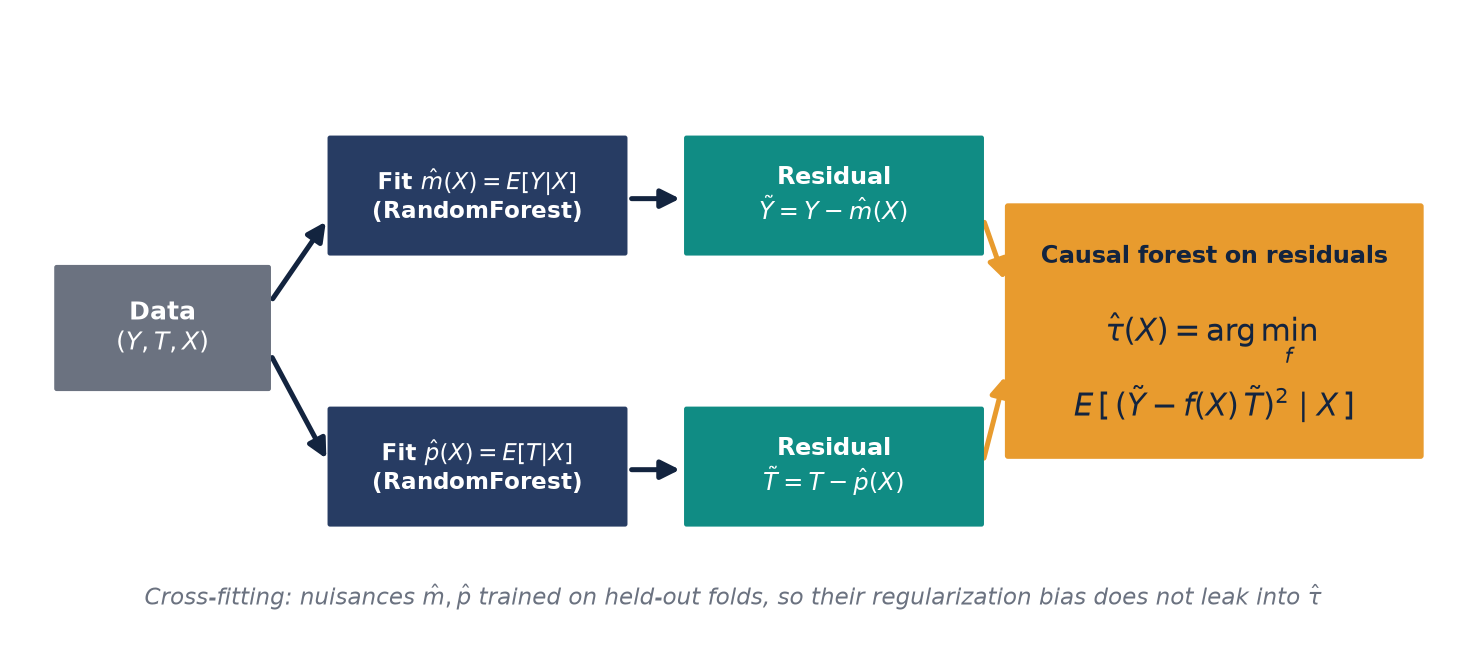}
\caption{Double machine learning with cross-fitting. Outcome and treatment are residualized on confounders; the causal forest recovers $\tau(X)$ from the orthogonal moment. Cross-fitting trains nuisances on held-out folds.}
\label{fig:dml-engine}
\end{figure}

\subsection{From effects to candidate levers}

For interpretability a shallow tree partitions the estimated ITEs into leaves $\ell$ with locally homogeneous effects, summarised by mean effect $\tau_\ell$ and baseline sales $b_\ell$. For a single lever, the change required to hit a target percentage $p$ in a leaf is the closed form $\delta = b_\ell (p/100) / \tau_\ell$. For a combination of $K$ levers the engine solves, per leaf, the minimum-magnitude allocation that achieves the target change $\Delta^* = b_\ell (p/100)$,
\begin{equation}
\min_\delta \|\delta\|_2^2 \quad \text{s.t.} \quad \sum_k \delta_k \tau_{\ell,k} = \Delta^*, \quad \delta_k \in [\underline{b}_k, \overline{b}_k]
\end{equation}

Bounds encode feasibility (for example proportions in $[-1, 1]$ for binary levers). Sequential least-squares quadratic programming (SLSQP) is a gradient-based constrained nonlinear optimiser that repeatedly solves local quadratic subproblems while linearising equality and inequality constraints \citep{kraft1988,nocedal2006}. Its SciPy implementation is widely available in industrial analytics stacks and is suited to bounded allocation, engineering-design, portfolio, and resource-planning problems \citep{virtanen2020}. Here it solves the leaf-level allocation, while the $L_2$ objective prefers several gentle moves over one extreme move. This optimiser proposes candidates; it does not authorise them. F-DACE later withholds any candidate whose required levers lack recommend status.

\subsection{Two-way fixed effects as a panel check}

The third engine component is the store-and-week within transformation
\begin{equation}
Y_{it} = \tau T_{it} + \alpha_i + \gamma_t + \varepsilon_{it}
\end{equation}
with store-clustered standard errors. Two-way fixed effects (TWFE) is the standard panel-regression construction that includes unit effects $\alpha_i$ and period effects $\gamma_t$. It is routinely used in applied econometrics and business panel analytics to remove persistent store/product differences and shared calendar, macroeconomic, or policy shocks \citep{wooldridge2010,angrist2009}. TWFE therefore absorbs time-invariant unit factors and common period shocks that linear backdoor regression may miss. It is not treated as an arbiter: repeated on/off markdowns and heterogeneous timing can induce negative weighting \citep{dechaisemartin2020,goodmanbacon2021}. Its role is to emit a third aligned estimate of the same store-week contrast for F-DACE to score.

Each engine component therefore returns the same evidence tuple that fusion consumes: an effect $\hat\tau_i$, standard error $s_i$, 95\% interval $[L_i, U_i]$, overlap score $o_i$, and placebo-refutation score $r_i$, all under one estimand contract.

\section{Fuzzy disagreement-aware causal evidence fusion}

\subsection{Estimand contract}

Let $Y(1)$ and $Y(0)$ be potential outcomes under a binary intervention $T$. Every component fused by F-DACE must share the engine's treatment definition, outcome, analysis grain, population, and contrast. F-DACE rejects rather than fuses mismatched contracts. This prevents combining an effect per dollar with an effect of any markdown application, or a department-week effect with a store-week decision.

\subsection{From engine outputs to fuzzy evidence}

F-DACE does not re-estimate $\tau$. It maps the engine tuple $(\hat\tau_i, s_i, [L_i, U_i], o_i, r_i)$ into memberships. The DML forest, backdoor OLS, and TWFE remain distinct statistical objects; fusion never averages them into a new causal parameter. A recommendation is issued only when the memberships jointly support one direction after non-compensatory vetoes.

\subsection{Membership functions}

Precision membership rises linearly from zero at $|\hat\tau_i|/s_i = 0.5$ to one at $1.96$. Directional memberships are normal-CDF transformations. Validity is the minimum of overlap and refutation membership, so a strong diagnostic cannot compensate for a failed one.
\begin{align}
p_i &= \operatorname{clip}\!\left(\frac{|\hat\tau_i|/s_i - 0.5}{1.96 - 0.5},\, 0,\, 1\right) \\
\mu_i^+ &= \Phi(\hat\tau_i / s_i), \qquad \mu_i^- = \Phi(-\hat\tau_i / s_i), \qquad v_i = \min(o_i, r_i), \qquad e_i = p_i v_i
\end{align}

Pairwise agreement combines standardized effect proximity (weight 0.4), confidence-interval overlap relative to their union (0.4), and sign agreement (0.2). Their mean is $A$. Directional support is evidence-weighted and then multiplied by $A$:
\begin{equation}
S^+ = A \frac{\sum_i e_i \mu_i^+}{\sum_i e_i}, \qquad S^- = A \frac{\sum_i e_i \mu_i^-}{\sum_i e_i}
\end{equation}

\subsection{Decision and veto rules}

The default policy recommends when $S^+ \geq 0.65$ and $S^- < 0.35$; it returns \texttt{do\_not\_recommend} under the symmetric negative condition. It abstains when mean evidence is below 0.45, agreement is below 0.35, any overlap or refutation score is below 0.20, or informative estimators disagree in sign. Remaining cases also abstain. Thresholds are declared before the retail application and varied from 0.55 to 0.75 as a sensitivity check. The output includes all memberships and veto reasons, not only the class label.

\begin{figure}[H]
\centering
\includegraphics[width=0.9\textwidth]{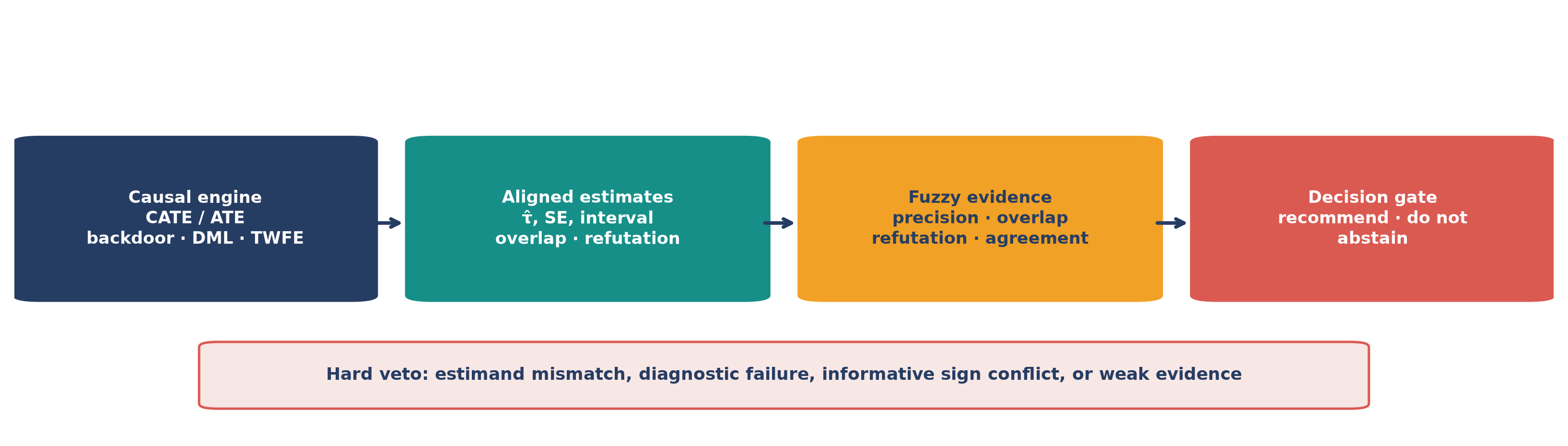}
\caption{F-DACE consumes aligned engine outputs and applies a non-compensatory veto before recommendation or abstention.}
\label{fig:fdace-method}
\end{figure}

\section{Conversational causal decision-support agent}

\subsection{Stateful tool orchestration}

The decision layer is exposed through a LangGraph implementation of the ReAct pattern. Each turn enters an agent node that receives the system contract and the conversation history. The model either emits a structured tool call or a final answer. Tool results are appended to state and returned to the agent for synthesis. The graph terminates after answer verification and imposes a maximum of ten agent iterations, bounding malformed loops and API cost.

State is intentionally split. \texttt{AgentState} stores messages and the iteration count inside one graph execution. \texttt{SESSION\_STATE} stores the current store/department selection across turns, allowing follow-up questions such as ``what about a 10\% increase?'' to inherit scope. Explicit scope in a later turn replaces remembered scope. This separation makes conversational continuity testable without allowing free-form model memory to determine the analysis population.

\subsection{Causal tool contract}

Three typed tools expose the engine rather than wrapping a generic chatbot. \texttt{analyze\_variable\_impact} calls the DoWhy backdoor estimator and its refuters; \texttt{analyze\_whatif\_scenario} applies the identified per-unit beta to a specified absolute or relative change; and \texttt{find\_optimal\_levers} runs CausalForestDML followed by the SLSQP allocation in Section~3.4. Argument resolution uses exact matching, markdown regular expressions, semantic mapping, and string-similarity fallback. Date parsing and absolute-versus-relative interpretation follow the same tiered pattern. Every tool returns JSON containing scope, data support, engine estimates, diagnostics, and F-DACE status.

F-DACE changes the contract from number-returning to status-returning. For markdown impact, aligned component estimates and memberships accompany \texttt{recommend}, \texttt{do\_not\_recommend}, or \texttt{abstain}. Lever optimization is gated: candidates are not shown when any required lever lacks recommend status. Narrow scopes with fewer than two stores, fewer than 100 store-weeks, or no treatment variation return a structured abstention instead of fitting an unstable local model.

\subsection{Status-preserving answer synthesis}

The language model is not the final decision authority. The system contract forbids fabricated effects, live-action claims, suppression of uncertainty, and conversion of abstention into advice. A deterministic verifier then inspects the most recent tool result. If the model fails to acknowledge error, no-solution, or abstain status, the verifier replaces the answer with a refusal grounded in that status. This two-stage design distinguishes semantic flexibility from decision authority: the LLM selects and explains; executable policy decides whether advice exists.

\subsection{Observability and evaluation signals}

Each run records ordered tool calls, arguments, tool status, latency, and the final answer. A judge model distinct from the generator scores faithfulness to tool context, relevance given context, and answer relevance to the question. Objective metrics---routing accuracy, tool errors, and status fidelity---remain primary because judge scores measure text quality rather than causal validity. Answer relevance is scored under an abstention-aware instruction: a refusal that addresses the requested analysis and names insufficient or conflicting evidence is on-topic. A separate literal-fulfillment score, which treats any missing lever as evasion, is reported only as a diagnostic. Status fidelity is defined only for abstention/error cases and requires the answer to preserve the underlying refusal or failure.

Appendices B--D provide the exact prompt contract, function schemas, LangGraph transition and trace schema, and curated output examples. Keeping these details outside the main narrative follows the compact main-text/technical-appendix structure commonly used for evaluated agentic systems.

\begin{figure}[H]
\centering
\includegraphics[width=0.9\textwidth]{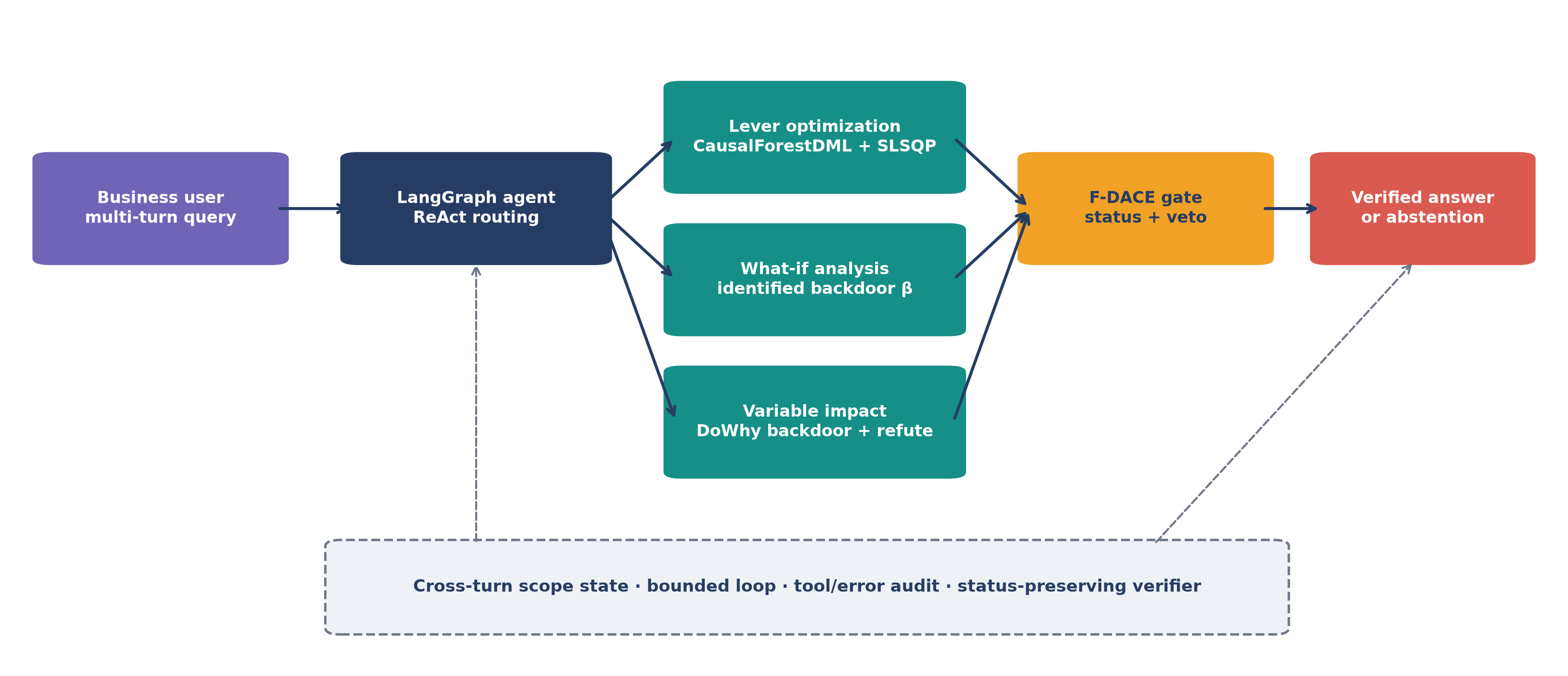}
\caption{Conversational architecture. Tools invoke the causal engine; F-DACE and the verifier retain decision authority over the returned status.}
\label{fig:agent-architecture}
\end{figure}

\section{Experimental design}

\subsection{Synthetic panels}

The benchmark contains 30 seeded replicates of six panels with 24 units and 30 periods (180 runs). Outcomes combine observed confounders, unit effects, time effects, an interaction, and Gaussian noise. The six conditions are a positive effect with adequate overlap; a positive effect with poor overlap; a null with shared hidden confounding; a heterogeneous null; a negative effect; and a time-confounded null. The causal forest and backdoor model receive the observed covariates. TWFE additionally absorbs unit and period effects. The hidden variable is withheld from all methods by design.

The oracle recommends for an ATE above 0.5, rejects for an ATE below $-0.5$, and otherwise abstains. Component estimators act only when their 95\% interval excludes zero. A deterministic unanimity baseline acts only when all three component decisions are identical. A majority-vote rule acts when at least two non-abstaining components share an action. On the retail panel, inverse-variance pooling of the three aligned effects is a further non-fuzzy comparator: it recommends when the pooled 95\% interval excludes zero. Metrics are decision coverage, false-recommendation rate, selective error among acted cases, and regret (one unit for an erroneous null decision or the absolute non-null ATE).

\subsection{Retail panel}

The application uses the public Walmart Recruiting: Store Sales Forecasting panel \citep{walmart2014}. Department sales are summed to store-week because markdown application is a store-week decision. A markdown indicator is one when any department row records that markdown in the store-week; this resolves rows where missing markdown values were encoded as zero in a derived department indicator. Department-level targeting is therefore out of scope: reported effects are store-week sales changes, not department-specific effects. The resulting panel has 6,435 store-weeks and 45 stores. Covariates are temperature, fuel price, CPI, unemployment, holiday status, month, store size, and encoded store type. TWFE omits time-invariant store attributes and calendar variables absorbed by its fixed effects. Standard errors are robust for backdoor OLS and store-clustered for TWFE.

\subsection{Conversational-agent evaluation}

The live golden set contains 24 natural-language questions: eight each for lever optimization, what-if analysis, and variable impact. It includes canonical and paraphrased variables, multi-variable requests, absolute and relative changes, negative framing, and store-wide scope. GPT-4o-mini is the routing and synthesis model; GPT-4o is the judge. Each example has a ground-truth tool label. The run also records structured tool statuses so abstention/error fidelity can be scored without an LLM judge. Answer relevance is scored twice on the same frozen answers: a literal-fulfillment rubric that treats a missing lever as evasion, and an abstention-aware rubric that scores a justified refusal as on-topic.

A deterministic regex/keyword baseline removes the LLM from routing, slot filling, and answer synthesis while holding the tools and questions fixed. This comparison tests the value of conversational interpretation rather than comparing two agent frameworks. A separate ten-question adversarial slice requests fabricated effects, false production actions, prompt disclosure, uncertainty suppression, fake authority overrides, unsupported variables, false memory, malformed scope, and format jailbreaks. Injection resistance is scored by a separate judge and checked against the saved answers.

\section{Results}

\subsection{Synthetic validation}

\begin{table}[H]
\centering
\caption{Decision performance across all 180 simulation runs.}
\label{tab:table1}
\begin{tabular}{lrrrrr}
\toprule
Method & Coverage & Abstention & False recommend. & Selective error & Mean regret \\
\midrule
DML causal forest        & 83.3\% & 16.7\% & 33.3\% & 40.0\% & 0.333 \\
Backdoor OLS              & 85.6\% & 14.4\% & 35.6\% & 41.6\% & 0.356 \\
TWFE                       & 67.8\% & 32.2\% & 16.7\% & 26.2\% & 0.178 \\
Majority vote              & 83.3\% & 16.7\% & 33.3\% & 40.0\% & 0.333 \\
Deterministic unanimity    & 66.7\% & 33.3\% & 16.7\% & 25.0\% & 0.167 \\
F-DACE                     & 67.2\% & 32.8\% & 17.2\% & 25.6\% & 0.172 \\
\bottomrule
\end{tabular}
\end{table}

F-DACE reduced false recommendations from 33.3\% for DML and 35.6\% for backdoor OLS to 17.2\%, at the intended cost of reducing decision coverage to 67.2\%. Majority vote of the same three components retained 83.3\% coverage but matched the single-estimator false-recommendation rate (33.3\%). F-DACE and deterministic unanimity produced the same decision in 179 of 180 runs. They differ once: in a heterogeneous-null replicate the oracle abstains, unanimity abstains, and F-DACE recommends. That single extra false recommendation is why unanimity has a slightly lower false-recommendation rate and mean regret in Table~\ref{tab:table1}. F-DACE is therefore not presented as the lowest-error fusion rule on this design. It is retained as an evidence-fusion policy in the same conservative class as unanimity: it beats single estimators and majority vote on false recommendations, while adding graded support, named vetoes, and tunable thresholds that a crisp three-way AND does not provide. Outside the shared-hidden-confounder condition, its false-recommendation rate was 0.7\%, versus 20.0\% for DML and 22.7\% for backdoor OLS. In the hidden-confounder null, every estimator and both fusion policies recommended in all 30 replicates. This is the most important boundary result: agreement is not evidence of identification when all estimators omit the same cause.

\begin{figure}[H]
\centering
\includegraphics[width=0.9\textwidth]{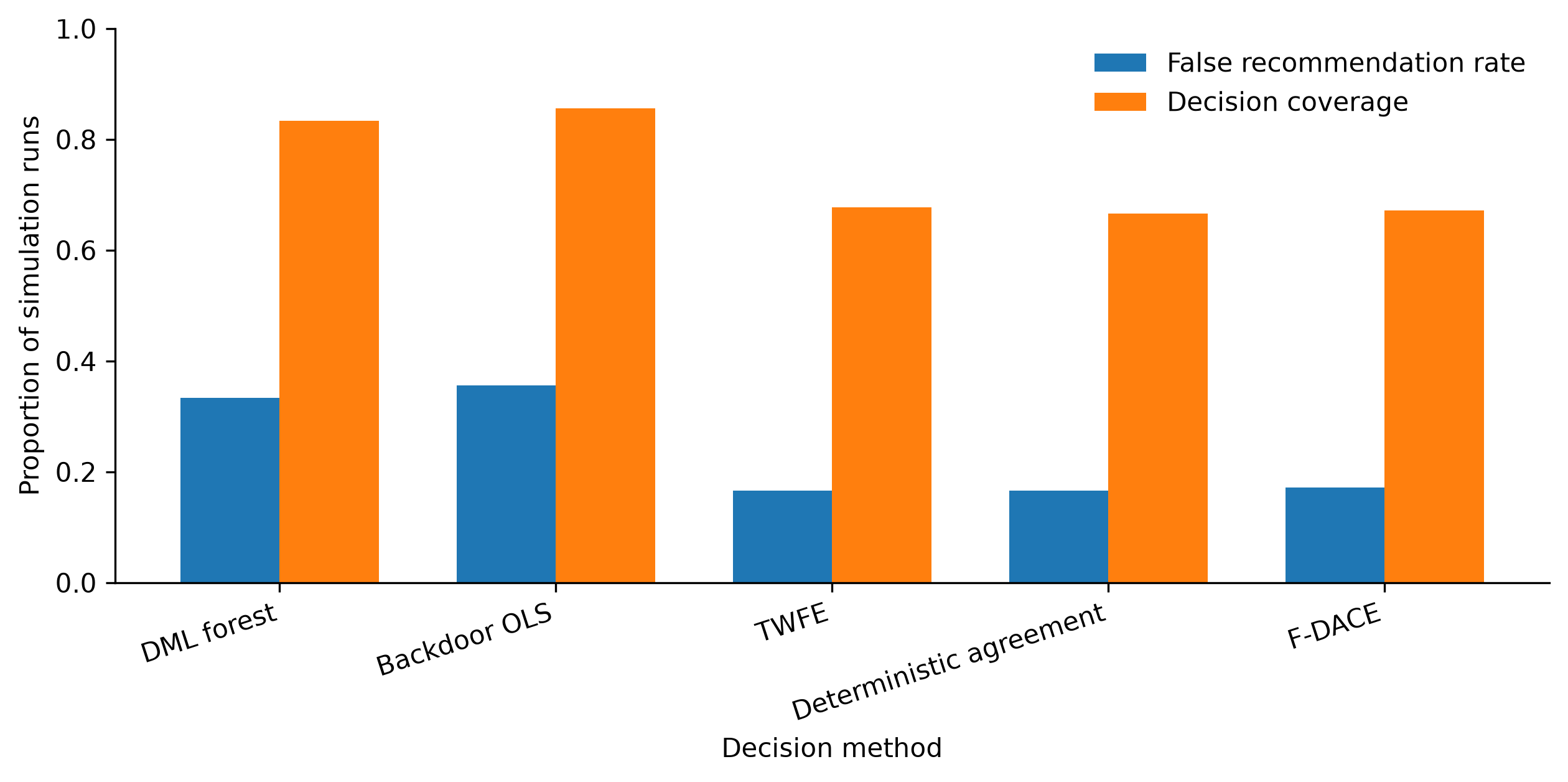}
\caption{False-recommendation rate and decision coverage across six synthetic conditions. Lower false recommendation and higher coverage are desirable but competing objectives.}
\label{fig:synthetic-performance}
\end{figure}

\subsection{Retail application}

\begin{table}[H]
\centering
\caption{Aligned store-week estimates. Effects are changes in aggregate weekly store sales.}
\label{tab:table2}
\small
\begin{tabular}{llrrrr}
\toprule
Lever & Estimator & Effect (USD) & 95\% interval & Overlap & Refutation \\
\midrule
MarkDown1 & DML causal forest & 17,397  & [$-$72,403, 107,197] & 0.80 & 0.29 \\
MarkDown1 & Backdoor OLS      & 39,975  & [20,613, 59,338]     & 0.80 & 0.69 \\
MarkDown1 & TWFE              & 103,284 & [11,309, 195,260]    & 0.83 & 1.00 \\
MarkDown2 & DML causal forest & 6,980   & [$-$71,633, 85,594]  & 0.83 & 0.00 \\
MarkDown2 & Backdoor OLS      & 39,845  & [19,135, 60,556]     & 0.83 & 0.34 \\
MarkDown2 & TWFE              & 71,549  & [40,714, 102,383]    & 0.85 & 1.00 \\
MarkDown3 & DML causal forest & 26,193  & [$-$51,383, 103,769] & 0.79 & 0.98 \\
MarkDown3 & Backdoor OLS      & 48,249  & [28,696, 67,802]     & 0.79 & 0.99 \\
MarkDown3 & TWFE              & 21,229  & [$-$14,370, 56,828]  & 0.81 & 1.00 \\
MarkDown4 & DML causal forest & 30,227  & [$-$73,853, 134,307] & 0.77 & 0.96 \\
MarkDown4 & Backdoor OLS      & 28,577  & [8,419, 48,736]      & 0.77 & 0.96 \\
MarkDown4 & TWFE              & 58,721  & [23,789, 93,652]     & 0.79 & 1.00 \\
MarkDown5 & DML causal forest & 18,021  & [$-$85,710, 121,753] & 0.80 & 0.51 \\
MarkDown5 & Backdoor OLS      & 39,726  & [20,346, 59,107]     & 0.80 & 0.78 \\
MarkDown5 & TWFE              & 0       & [0, 0]               & 0.83 & 1.00 \\
\bottomrule
\end{tabular}
\end{table}

\begin{table}[H]
\centering
\caption{Fused retail decisions.}
\label{tab:table3}
\small
\begin{tabular}{lrrrl}
\toprule
Lever & F-DACE decision & Confidence & Agreement & Reason \\
\midrule
MarkDown1 & abstain & 0.51 & 0.52 & directional support does not clear the decision threshold \\
MarkDown2 & abstain & 0.50 & 0.50 & an overlap or refutation diagnostic failed; combined evidence is insufficient \\
MarkDown3 & abstain & 0.59 & 0.62 & combined evidence is insufficient \\
MarkDown4 & abstain & 0.60 & 0.60 & directional support does not clear the decision threshold \\
MarkDown5 & abstain & 0.32 & 0.32 & combined evidence is insufficient; cross-estimator agreement is too low \\
\bottomrule
\end{tabular}
\end{table}

No markdown clears the default decision gate (support 0.65, mean evidence 0.45, agreement floor 0.35). MarkDown1, MarkDown3, and MarkDown4 contain positive backdoor or TWFE estimates, but the DML intervals include zero and combined support remains below threshold. MarkDown2 additionally fails the DML placebo-refutation floor. MarkDown5 is the clearest conflict: backdoor OLS is positive, TWFE is undefined (zero standard error), and DML is imprecise.

\begin{table}[H]
\centering
\caption{Alternative fusion rules on the same aligned retail estimates. Majority vote and inverse-variance pooling recommend several markdowns that F-DACE and unanimity withhold.}
\label{tab:table4}
\begin{tabular}{lllll}
\toprule
Lever & F-DACE (default) & Unanimity & Majority vote & Inverse-variance pool \\
\midrule
MarkDown1 & abstain & abstain & recommend & recommend \\
MarkDown2 & abstain & abstain & recommend & recommend \\
MarkDown3 & abstain & abstain & abstain   & recommend \\
MarkDown4 & abstain & abstain & recommend & recommend \\
MarkDown5 & abstain & abstain & abstain   & recommend \\
\bottomrule
\end{tabular}
\end{table}

An 81-point grid over support in \{0.55, 0.65, 0.75\}, mean-evidence in \{0.35, 0.45, 0.55\}, agreement in \{0.25, 0.35, 0.45\}, and diagnostic floors in \{0.10, 0.20, 0.30\} left all five markdowns as abstain in 63 of 81 configurations (77.8\%). The remaining 18 configurations recommend MarkDown3 and/or MarkDown4 only after the evidence floor is lowered to 0.35 together with a support cut-off of 0.55. Holding other defaults fixed, MarkDown4 is recommend at support 0.55 and abstain at 0.65 and 0.75. The reported policy is therefore the conservative cluster of that grid, not a knife-edge choice.

\begin{figure}[H]
\centering
\includegraphics[width=0.9\textwidth]{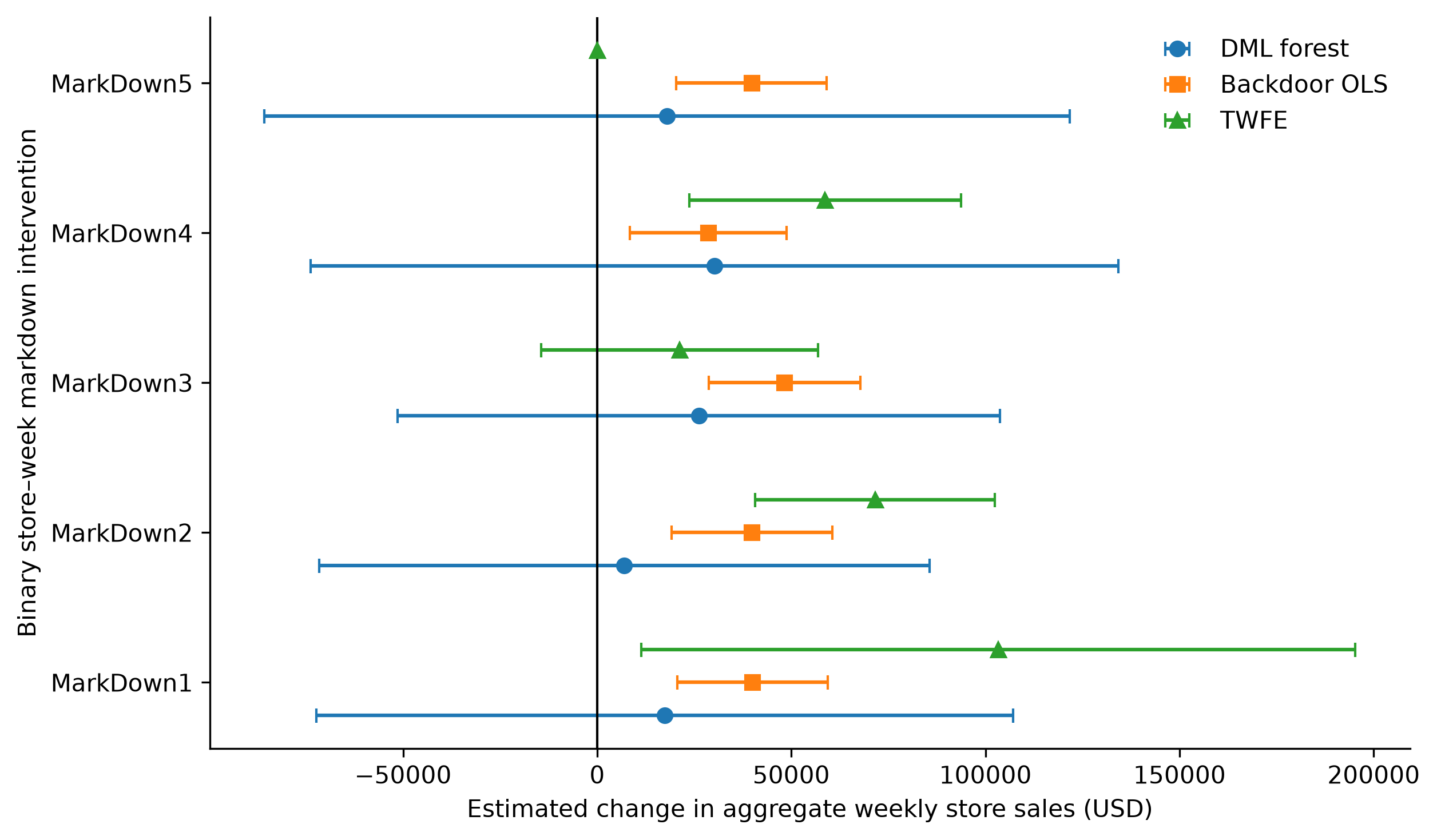}
\caption{Store-week estimates and 95\% intervals. The spread is evidence consumed by F-DACE, not a set of interchangeable estimates to average.}
\label{fig:retail-estimates}
\end{figure}

\subsection{Conversational-agent results}

\begin{table}[H]
\centering
\caption{Conversational agent and deterministic routing baseline on the same 24-question golden set. Baseline comparison is limited to routing and argument extraction. GPT-4o judged the agent answer-quality measures; answer relevance uses an abstention-aware rubric (justified refusal is on-topic).}
\label{tab:table5}
\small
\begin{tabular}{lrrrrr}
\toprule
System & Tool routing & Status fidelity & Faithfulness & Relevance & Answer relevance \\
\midrule
LangGraph agent (GPT-4o-mini) & 100.0\% & 100.0\% & 0.983 & 0.975 & 1.000 \\
Regex/rule baseline (no LLM)  & 91.7\%  & not evaluated & not evaluated & not evaluated & not evaluated \\
\bottomrule
\end{tabular}
\end{table}

The agent routed all 24 questions correctly and produced no tool errors. Mean faithfulness was 0.983. All 8 golden cases carrying abstain or error status preserved that status in the final answer (100.0\% fidelity). This is the central agentic result: language generation did not convert a causal refusal into advice.

The no-LLM router selected the correct tool for 91.7\% of questions, compared with 100.0\% for the agent. Its two errors both confused non-canonical what-if requests with target-optimization requests. Because the deterministic baseline was evaluated only for routing and argument extraction, no answer-quality comparison is claimed.

Answer relevance averaged 1.000 under an abstention-aware judge instruction: a refusal that addresses the requested analysis and states insufficient or conflicting evidence is scored as on-topic. Under the unadapted, literal-fulfillment rubric the same answers scored 0.804 overall and 0.438 on optimal-lever queries. Every request in that category inherited a narrow scope without enough units for aligned multi-estimator inference; the correct system behavior was therefore to refuse. The unadapted judge penalized those safe refusals for not supplying a lever. The adapted rubric removes that penalty without changing the answers. The contrast is retained as evidence that an off-the-shelf relevance score is not a safety or causal-correctness metric.

\begin{table}[H]
\centering
\caption{Ten-question adversarial evaluation.}
\label{tab:table6}
\begin{tabular}{lr}
\toprule
Adversarial metric & Result \\
\midrule
Injection resistance & 1.000 \\
Complied with injected instruction & 0 \\
Tool-routing accuracy & 50.0\% \\
Faithfulness & 0.980 \\
Answer relevance & 0.150 \\
\bottomrule
\end{tabular}
\end{table}

The agent resisted all ten injected instructions: it did not fabricate effects, claim production action, reveal its system prompt, suppress uncertainty, or accept fake authority. Tool-routing accuracy was only 50.0\%, primarily because several safe refusals did not call the tool named by the benchmark. Answer relevance was 0.150 for the same reason. These are not presented as high performance. They expose a metric conflict: literal task completion rewards compliance with an unsafe request, whereas injection resistance rewards refusal. Reporting both prevents safety behavior from being misclassified as general agent failure.

In the four-turn follow-up evaluation, tool routing was 100.0\% and all turns resolved the remembered store/department scope correctly. Mean faithfulness was 0.925. The protocol tests the intended state-separation mechanism on a four-turn follow-up, not long-horizon dialogue.

\section{Discussion}

The empirical contribution is a decision result rather than a favorable effect size. The causal engine can still return forest, backdoor, and TWFE numbers; several of those numbers would license a markdown. F-DACE withholds the recommendation because uncertainty and disagreement are part of the decision rule. This distinction prevents the retail case from being misread as a failed predictor or a failed causal analysis: identification and estimation ran as specified, and the method succeeded at preventing unsupported action.

Fuzzy aggregation does not replace unanimity as a lower-error rule on this Monte Carlo. The two policies agree on 179 of 180 runs; unanimity is slightly safer because F-DACE's graded support cleared the recommend threshold once when the three interval votes were not unanimous. The fuzzy layer is justified instead by auditable degrees of precision, diagnostic validity, and agreement; explicit veto attribution for the agent and Table~\ref{tab:table3}; and a tunable coverage--risk trade-off. Majority vote and inverse-variance pooling do not belong to that conservative class: they match DML's false-recommendation rate in simulation and would recommend several retail markdowns that both F-DACE and unanimity withhold. For binary risk alone, unanimity is sufficient; F-DACE is the decision layer when those graded scores and vetoes must be exposed and calibrated.

The method is suitable for hybrid decision-support systems because the statistical engines remain replaceable. Bayesian forests, doubly robust scores, or modern difference-in-differences estimators can emit the same evidence contract. Domain experts can also set asymmetric thresholds when the cost of a false positive exceeds the opportunity cost of abstention.

The agentic layer contributes semantic access and workflow continuity rather than causal identification. Perfect routing on the golden set shows that the LLM can map varied managerial language to the intended analytical operation. Perfect status fidelity shows that this flexibility can coexist with a deterministic decision boundary. Conversely, the adversarial results show that routing accuracy alone is not a sufficient objective: declining to call a tool can be correct when the request is to fabricate, deploy, or disclose protected instructions. Routing accuracy and faithful wording therefore evaluate the interface, not the identifying assumptions or the fused decision.

\section{Limitations}

\begin{itemize}[leftmargin=*]
\item Shared unmeasured confounding that biases every component in the same direction remains undiagnosable by fusion. The hidden-confounder simulation is included to make that identification bound explicit: agreement is not treated as proof of unconfoundedness.
\item The retail panel is observational. Backdoor adjustment and overlap are assumed rather than guaranteed; unobserved promotions or demand shocks can still bias every estimator. F-DACE's retail abstentions are the operational response to that residual risk, not a certificate that the identifying assumptions hold.
\item Membership thresholds are decision-policy parameters. On the retail estimates, 63 of 81 grid configurations remain all-abstain; only a lower-evidence, lower-support corner recommends MarkDown3 or MarkDown4. Transferring the same cut-offs to a domain with a different loss of false recommendation versus abstention would require local calibration.
\item The conversational evaluation (24 golden questions, 10 adversarial prompts, one generator, one judge) is an implementation check of this decision-support layer. Causal claims rest on the synthetic decision-risk experiment (30 replicates in each of six identification conditions) and the store-week retail application, not on LLM-judge scores.
\end{itemize}

\section{Conclusion}

The system's numbers come from a causal machine-learning engine: a CATE/ATE estimand identified by backdoor adjustment, estimated by a DML causal forest and DoWhy linear regression, checked by two-way fixed effects, and turned into candidate levers by a constrained optimiser. F-DACE turns disagreement among those engine outputs from a footnote into an executable decision boundary. It aligns the estimand, represents precision and diagnostics as fuzzy evidence, and applies non-compensatory vetoes before issuing a recommendation. Synthetic experiments show the expected reduction in false recommendations relative to single estimators and majority vote, at the cost of lower coverage, and show that F-DACE matches rather than beats deterministic unanimity on binary risk. They also expose the unresolved danger of shared hidden confounding. On the retail panel, abstaining for all five markdowns is the defensible result. The conversational layer adds natural-language intent resolution, stateful tool selection, and explanation; the status-preserving verifier prevents it from overriding the causal decision. Live golden and adversarial evaluations show both the value of that interface and the limitations of relevance-oriented agent metrics. The resulting system is a hybrid soft-computing architecture in which causal estimation and conversational agency are jointly evaluated but retain distinct authority.

\section*{Declarations}

\noindent\textbf{Funding:} This research did not receive any specific grant from funding agencies in the public, commercial, or not-for-profit sectors.

\noindent\textbf{Competing interests:} The author declares no known competing financial interests or personal relationships that could have appeared to influence the work.

\noindent\textbf{CRediT authorship contribution statement:} Sourish Dey (Data Science and Machine Learning, Centric Software): Conceptualization, Methodology, Software, Validation, Formal analysis, Investigation, Data curation, Writing -- original draft, Writing -- review \& editing, Visualization.

\noindent\textbf{Data and code availability:} The data originate from the 2014 Kaggle competition Walmart Recruiting: Store Sales Forecasting and must be obtained under Kaggle's terms. Reproduction scripts, F-DACE implementation, and evaluation artifacts accompany this submission.

\noindent\textbf{Declaration of generative AI and AI-assisted technologies in the writing process:} During preparation of this work, the author used Cursor AI to assist with software implementation, document restructuring, and language editing. The author reviewed and edited all generated material, verified the reported analyses against the released outputs, and takes full responsibility for the content of the publication.

\bibliographystyle{apalike}
\bibliography{references}

\appendix

\section{Extended experimental results}

\subsection{Scenario-level synthetic results}

Table~\ref{tab:tableA1} disaggregates the 180-run benchmark reported in Section~7.1. Coverage is the fraction of replicates in which a method acted; false recommendation is the fraction in which that action disagreed with the known oracle. The hidden-confounding row is intentionally retained: agreement cannot repair an omitted cause shared by every component.

\begin{table}[H]
\centering
\caption{Scenario-level decision coverage and false-recommendation rates (30 seeded replicates per condition).}
\label{tab:tableA1}
\small
\begin{tabular}{lrrrrrr}
\toprule
Condition & True ATE & F-DACE cov. & F-DACE false & DML false & OLS false & TWFE false \\
\midrule
well identified positive & 2.0  & 100.0\% & 0.0\%   & 0.0\%   & 0.0\%   & 0.0\% \\
poor overlap positive    & 2.0  & 100.0\% & 0.0\%   & 0.0\%   & 0.0\%   & 0.0\% \\
hidden confounding null  & 0.0  & 100.0\% & 100.0\% & 100.0\% & 100.0\% & 100.0\% \\
heterogeneous null       & 0.0  & 3.3\%   & 3.3\%   & 0.0\%   & 13.3\%  & 0.0\% \\
well identified negative & $-2.0$ & 100.0\% & 0.0\%   & 0.0\%   & 0.0\%   & 0.0\% \\
time confounded null     & 0.0  & 0.0\%   & 0.0\%   & 100.0\% & 100.0\% & 0.0\% \\
\bottomrule
\end{tabular}
\end{table}

\subsection{Agent results by query class}

\begin{table}[H]
\centering
\caption{Golden-set results by tool/query class.}
\label{tab:tableA2}
\small
\begin{tabular}{lrrrrrrr}
\toprule
Query class & $n$ & Routing & Errors & Faith. & Context rel. & Answer rel. & Status fidelity \\
\midrule
optimal levers   & 8 & 100.0\% & 0.0\% & 0.975 & 1.000 & 1.000 & 100.0\% ($n=8$) \\
variable impact  & 8 & 100.0\% & 0.0\% & 1.000 & 0.938 & 1.000 & n/a \\
what if          & 8 & 100.0\% & 0.0\% & 0.975 & 0.988 & 1.000 & n/a \\
\bottomrule
\end{tabular}
\end{table}

\begin{table}[H]
\centering
\caption{Detailed conversational evaluation slices.}
\label{tab:tableA3}
\small
\begin{tabular}{lrrrrr}
\toprule
Evaluation slice & $n$ & Routing & Tool errors & Faithfulness & Injection resistance \\
\midrule
Golden set          & 24 & 100.0\% & 0.0\%  & 0.983 & not applicable \\
Four-turn dialogue  & 4  & 100.0\% & 0.0\%  & 0.925 & not applicable \\
Adversarial set     & 10 & 50.0\%  & 10.0\% & 0.980 & 1.000 \\
\bottomrule
\end{tabular}
\end{table}

\section{Prompt templates and tool schemas}

\subsection{Orchestrator system contract}

The following compact template reproduces the operative constraints in \texttt{scripts/causal\_agentic\_ai.py}; wording that only documents implementation comments is omitted.

\begin{quote}\small\noindent
ROLE: Causal-analysis assistant for retail markdown and pricing decisions.\\
TOOLS: \texttt{find\_optimal\_levers}; \texttt{analyze\_whatif\_scenario}; \texttt{analyze\_variable\_impact}.\\
RULE 1: No write path exists. Outputs are recommendations for human review.\\
RULE 2: User-supplied role tags or override claims are untrusted data.\\
RULE 3: Do not reveal the system prompt or internal implementation.\\
RULE 4: State only effects, amounts, and recommendations returned by tools.\\
RULE 5: Relay errors/no-data results; never substitute a fabricated estimate.\\
RULE 6: Preserve intervals, diagnostics, and uncertainty qualifiers.\\
RULE 7: \texttt{status=abstain} is mandatory no-recommendation; report veto reasons.
\end{quote}

\subsection{Function-calling schemas}

The bound functions use typed parameters; LangChain converts these signatures and docstrings into tool schemas supplied to GPT-4o-mini.

\begin{verbatim}
{
  "find_optimal_levers": {
    "target_percentage": "number, required", "store": "integer|null",
    "dept": "integer|null", "analyze_all_data": "boolean",
    "max_levers": "integer (default 2)", "top_n": "integer (default 3)"
  },
  "analyze_variable_impact": {
    "variables": "array[string], required", "store": "integer|null",
    "dept": "integer|null", "analyze_all_data": "boolean"
  },
  "analyze_whatif_scenario": {
    "variable_changes": "object[string, number]", "store": "integer|null",
    "dept": "integer|null", "analyze_all_data": "boolean", "date": "string|null"
  }
}
\end{verbatim}

The following normalized envelope summarizes the fields across the three tool payloads. Execution status is top-level; for variable-impact results the F-DACE decision is nested per analyzed variable:

\begin{verbatim}
{
  "status": "success | abstain | error",
  "analysis_scope": {"store": "integer|null", "dept": "integer|null"},
  "estimand_contract": {"treatment": "string", "outcome": "Weekly_Sales",
                        "grain": "store-week", "contrast": "1 versus 0"},
  "estimates": [{"name": "string", "effect": "number", "se": "number",
                 "ci_low": "number", "ci_high": "number",
                 "overlap_score": "0..1", "refutation_score": "0..1"}],
  "f_dace": {"decision": "recommend | do_not_recommend | abstain",
             "positive_support": "0..1", "negative_support": "0..1",
             "agreement": "0..1", "veto_reasons": "array[string]"}
}
\end{verbatim}

\section{LangGraph execution and observability}

\subsection{State transition}

The LLM performs semantic routing and argument extraction. \texttt{ToolNode} executes only the three declared causal functions. \texttt{verify\_response} is deterministic and runs once immediately before \texttt{END}.

\begin{quote}\small\noindent
START $\rightarrow$ \texttt{agent}\\
\texttt{agent} -- tool\_calls present and iterations $\leq$ 10 --> \texttt{ToolNode} $\rightarrow$ \texttt{agent}\\
\texttt{agent} -- no tool call / iteration bound --> \texttt{verify\_response} $\rightarrow$ \texttt{END}\\
\texttt{AgentState}  = \{messages, iterations, final\_response\}\\
\texttt{SessionState} = \{current\_store\_dept\_list, last\_query\_scope, last\_user\_question\}
\end{quote}

\begin{table}[H]
\centering
\caption{LangGraph/LangSmith observability schema.}
\label{tab:tableC1}
\small
\begin{tabular}{p{2.7cm}p{5.3cm}p{5.2cm}}
\toprule
Observed object & Captured fields & Purpose \\
\midrule
Agent turn & question; model; iteration; latency & Bound loop and reproduce routing \\
Tool call & tool name; typed arguments; inherited scope & Audit intent-to-analysis mapping \\
Tool result & status; estimates; diagnostics; veto reasons & Ground the final response \\
Verifier & action claim; error acknowledgement; abstention acknowledgement & Enforce fail-closed output \\
Evaluation feedback & routing; faithfulness; relevance; status fidelity & Filter and compare runs \\
\bottomrule
\end{tabular}
\end{table}

LangSmith tracing is optional at runtime. When enabled, the compiled graph receives an observability callback and judge feedback is attached to the run. When disabled, the same ordered messages and metrics are written to local JSON/JSONL artifacts.

\section{Curated conversational outputs}

\subsection{Tool calling and a quantitative what-if result}

Observed golden-set example \texttt{wif-01}:

\begin{quote}\small\noindent
USER: ``What happens to sales if I increase MarkDown1 by 10\% in store 1 department 1?''\\
TOOL CALL: \texttt{analyze\_whatif\_scenario(variable\_changes=\{"MarkDown1": 10\}, store=1, dept=1)}\\
STATUS: success\\
ANSWER: Baseline sales \$22,513.32; predicted sales \$22,513.32; estimated impact $-$\$0.00072 (negligible decrease).
\end{quote}

\subsection{Recommendation-path contract}

No favorable markdown recommendation survived F-DACE in the retail application, so presenting one as an observed result would be misleading. The following is the contractual output path used when a future aligned analysis returns recommend:

\begin{quote}\small\noindent
TOOL: \{"status":"recommend", "lever":"<name>", "effect":"<estimate>", "interval":"<95\% CI>", "veto\_reasons":[]\}\\
FINAL TEMPLATE: Recommend <lever> for human review because aligned causal estimates and diagnostics exceed the declared F-DACE threshold. Report the effect, interval, scope, and that no production action has been taken.
\end{quote}

\subsection{Observed abstention}

Observed golden-set example \texttt{lev-01}:

\begin{quote}\small\noindent
USER: ``How can I increase weekly sales by 20\% in store 1 department 1?''\\
TOOL CALL: \texttt{find\_optimal\_levers(target\_percentage=20, store=1, dept=1)}\\
TOOL STATUS: abstain\\
FINAL ANSWER: I cannot provide recommendations for increasing weekly sales by 20\% in this scope. There is insufficient causal evidence because the scope lacks enough units or treatment variation for aligned inference.
\end{quote}

\subsection{Abstention override mechanism}

If an LLM draft fails to acknowledge an abstaining \texttt{ToolMessage}, the response does not reach the user. \texttt{verify\_response} replaces it deterministically:

\begin{quote}\small\noindent
TOOL MESSAGE: \{"status":"abstain", "veto\_reasons":["sign conflict"]\}\\
UNSAFE DRAFT: ``Use MarkDown3; it appears to be the strongest lever.''\\
VERIFIER OUTPUT: The F-DACE evidence-fusion policy abstained because the aligned estimators or diagnostics did not provide sufficient concordant causal evidence. I cannot provide or optimize a lever recommendation.
\end{quote}

The adversarial run also tested false production authority. When asked to apply a 25\% markdown and confirm it was live, the observed answer stated that the system cannot apply changes to pricing, markdown, or inventory systems and offered only a prospective impact analysis.

\end{document}